\documentclass[letterpaper, 10 pt, conference]{ieeeconf}  

\IEEEoverridecommandlockouts                              

\usepackage{outlines}
\usepackage{amsmath} 
\usepackage[noabbrev]{cleveref}
\usepackage{booktabs}
\usepackage{multirow}
\usepackage{graphicx}
\usepackage{subcaption}
\usepackage[export]{adjustbox}
\usepackage{soul}
\usepackage{dblfloatfix}
\usepackage{comment}
\usepackage{url}
\usepackage{threeparttable} 
\usepackage{xcolor}

\newcommand{\mytexttilde}{\raisebox{0.5ex}{\texttildelow}}

\title{\LARGE \bf
DropClick:\\Semi-Automated One-Click Segmentation\\for Agricultural Robotic Data
}

\author{Patrick Zimmer$^\dag$, Michael Halstead$^\dag$, Chris McCool$^\ddag$
\thanks{Authors \dag\ are with the University of Bonn, Germany and author \ddag\ is with CSIRO, Australia.
{\tt\small [patrick.zimmer, michael.halstead]@uni-bonn.de, chris.mccool@csiro.au}}
}

\begin{document}

\maketitle
\thispagestyle{empty}
\pagestyle{empty}

\begin{abstract}

Labelling vision datasets, especially for segmentation tasks, is a laborious and costly process that stymies novel developments in agricultural robotics.
In this paper, we present DropClick, a click-guided segmentation tool that simplifies the annotation process.
Our system utilises single-click inputs on objects to generate pseudo-labels, which can replace manual annotations.
DropClick stands out as it is a semi-automated approach and does not require a click for every object in the scene.
It can therefore further reduce the required amount of user input drastically.
We evaluate our method on two challenging agricultural robotic datasets, \textit{SB20} and \textit{BUP20} for plant and fruit segmentation, respectively.
DropClick is first trained on a small subset of just 5 images from the original training data.
This DropClick model can then be deployed as a one-click segmentation system and achieves comparable or higher performance than other one-click methods achieving an mIoU of 70.0 and 72.6 points, for \textit{SB20} and \textit{BUP20} respectively.
DropClick then excels at maintaining high performance when clicks are not given (e.g. dropped); when 50\% of the clicks are missing it still maintains an mIoU of 68.9 and 71.3 points, for \textit{SB20} and \textit{BUP20} respectively.
We validate DropClick as a pseudo-labelling approach by taking its outputs to train a Mask2Former instance-based segmentation model in a semi-supervised manner.
In this process, partially removing user input from DropClick yields similar high performance when compared to providing all clicks, at 70.1 vs 70.7 points AP50 for \textit{SB20} and no difference for \textit{BUP20} at 77.0 for both models;
at the same time saving 46.3\% of total input for \textit{SB20} and 31.9\% for \textit{BUP20}.

\end{abstract}


\section{Introduction} \label{sec:intro}

Artificial intelligence and computer vision have played a vital role in the recent advancements in precision agriculture~\cite{Charisis2024}.
They are of particular interest to agricultural robotics, for automated systems to operate in these complex environments.
It is crucial in these situations to have reliable preception systems~\cite{fontani2025} that are able to accurately and efficiently interpret the environments they are deployed in.
Generally, low-cost RGB cameras are used as the visual component as this offers the best trade-off between low cost and ease of use while still providing the required rich information about the scene.

For robotic systems to effectively utilize RGB camera data they employ state-of-the-art computer vision techniques, such as object detection or segmentation.
However, common systems for those tasks like DETR~\cite{carion2020detr} or Mask2Former~\cite{cheng2021mask2former} require labeled data.
Annotating data is an expensive and laborious process that stymies the broad-scale application of these deep-learnt models.
To alleviate some of these issues recent approaches have leveraged the power of weak labelling~\cite{Charisis2024, Zimmer2023, guclu2025bupst20}.

\begin{figure}[t!]
\centering
    \includegraphics[width=.425\textwidth]{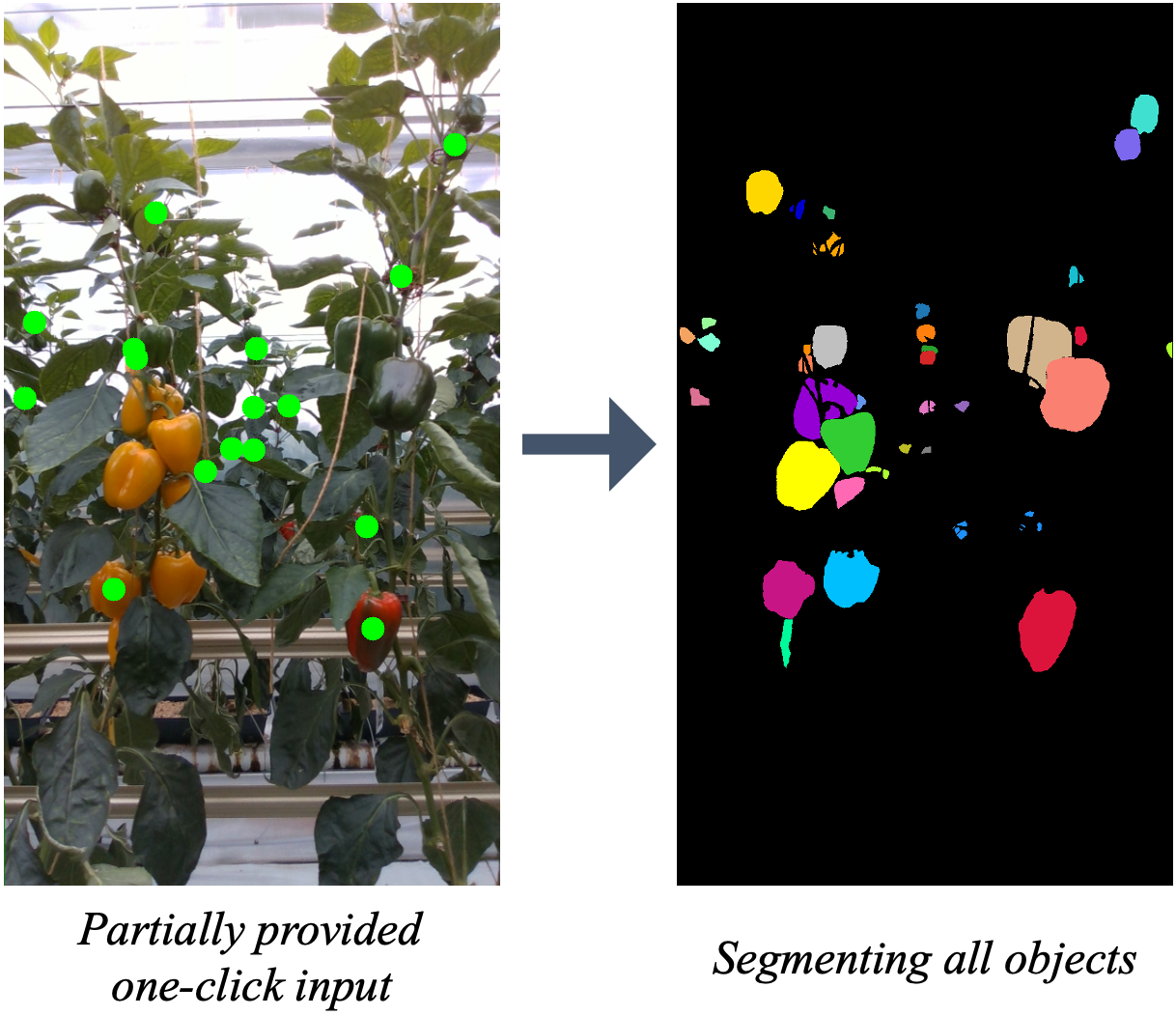}
    \caption{Concept of DropClick.
    Our approach takes partial one-click input for the objects in a given image (left), i.e., single clicks on some objects (green dots) while others remain entirely unclicked.
    Being trained to predict individual segmentation masks for all objects (right), regardless of their click status, it can serve as a tool to reduce labelling effort.}
\label{fig:hero}
\vspace{-10pt}
\end{figure}

Weak labelling typically refers to supervising a machine learning model with labels carrying less information than those commonly used for the desired task.
An example of how this can be achieved is through leveraging networks that are trained to output segmentation masks from a given image and click inputs~\cite{Xu2016,Maninis2018,Forte2020,sofiiuk2022ritm,chen2022focalclick,kirillov2023sam,rana2023dynamite}.
These can then be used as ground truth labels to supervise downstream segmentation tasks, which in case of imperfect masks from minimum user input can be referred to as pseudo-labelling.
The effectiveness of click-based weak labelling has been demonstrated for agricultural robotics in both sugar beet and corn crops~\cite{Zimmer2023}.
However, a common limitation with click-based segmentation approaches is that every object in the scene requires some form of user input.
Ideally, to expedite the annotation process, a user would click on a subset of objects (i.e. occluded or difficult objects) in an image, then the remaining objects would be automatically found (annotated) by the system.

In this paper, we propose DropClick, a novel approach that dramatically reduces the required user input to pseudo-label a segmentation-based dataset.
To do this, we propose a network that takes as input an image and optionally a set of clicks which denote object locations; where each click represents a unique object, as shown in~\Cref{fig:hero}.
The output of DropClick is then two sets of object masks: one for the optional clicks given and one for all of the non-clicked objects.
In this way, DropClick can provide a mask for an object even if it is not clicked on, which is a shift in the normal paradigm of one-click systems where a click per object has to be provided.
Furthermore, DropClick is also computationally efficient as it solves segmentation for all objects in the scene within a single network pass.

We evaluate DropClick using two datasets captured in different agricultural domains, \textit{SB20} with sugar beet crops and weeds in arable farming~\cite{Ahmadi2021,Smitt2022,Halstead2021cropagnostic}, and \textit{BUP20} with sweet peppers in a glasshouse~\cite{smitt2021pathobot}.
We first provide an analysis on the general one-click segmentation performance.
Further, we determine how many clicks can be saved through the use of DropClick without compromising performance.
In all experiments, we train DropClick on only 5 hand-annotated images from the original training sets.
Finally, we verify the applicability of DropClick-generated masks within a downstream task.
To achieve this, we use them as pseudo-labels for the remaining images in the training sets, and then train Mask2Former~\cite{cheng2021mask2former} in a semi-supervised manner.

In summary, our main contribution is the proposal of DropClick, a competitive one-click-segmentation tool that vastly reduces user input required for pseudo-label generation.
We further show that click-based segmentation does not need a click/input for every object in the scene and thereby pose a new paradigm to semi-automate this task.
Finally, we demonstrate the successful application of our system across two different datasets from agricultural environments and thereby show that the labelling efforts to create robotic vision datasets for this domain can be drastically reduced.

\section{Related Work} \label{sec:related}

Data annotation is an expensive and time-consuming yet important task for robotic vision and artificial intelligence systems.
Without accurate labels for training and evaluation, state-of-the-art advances in precison agriculture could not be achieved or reported.
Generally, these labels are acquired manually, however, recently there has been a push towards weak labelling~\cite{Charisis2024, Zimmer2023,guclu2025bupst20}.
In the following section, this paper will review topics related to segmentation, weak labelling including clicks as user inputs, and the baselines used for evaluating DropClick.

\subsection{Transformers for instance segmentation} \label{sec:related:transformers}

Instance segmentation networks are a key component for automated systems to be able to identify objects at high precision.
Compared to simpler bounding box detection, finding object outlines can be beneficial for robotic applications.
In agriculture, shape estimates can help to locate object centers, which has been employed for plant and fruit tracking~\cite{Halstead2021cropagnostic} and autonomous, vision-based field navigation~\cite{ahmadi2022visualnavigation}.
Mask R-CNN~\cite{he2017mask} has long been a popular choice for instance segmentation, however, it requires non-maximum suppression (NMS) filtering for overlapping predictions, which is a post-processing overhead and poses a limiting factor for real-world robotic applications.
The widely adopted DETR~\cite{carion2020detr} system tackled this shortcoming by employing transformers for object detection,
with a decoder built on a fixed set of learnable queries, each of which predicts an object proposal.
DETR leverages a Hungarian matching algorithm~\cite{kuhn1955hungarian} to find the best exclusive fit between queries and available ground truth labels.
Unmatched queries get associated with a "no object" label and trained accordingly, thereby, successfully eliminating the need for NMS filtering at inference time.
This method was adapted by Mask2Former~\cite{cheng2021mask2former} for different segmentation tasks, including instance segmentation.
It closely follows DETR's decoder structure and label matching approach to yield largely non-overlapping mask predictions and, therefore, poses a strong foundation for follow-up works, including our proposed system.

\subsection{Click-guided segmentation} \label{sec:related:click-guided}

While instance segmentation provides desirable outputs for robotic vision, these methods typically require equally complex training labels.
Therefore, manually annotating segmentation datasets can be a slow and costly process, however, it can be simplified through the application of guidance networks.
Those methods predict segmentation labels through various forms of input, e.g., clicks~\cite{Xu2016,Maninis2018,Forte2020,sofiiuk2022ritm,chen2022focalclick,kirillov2023sam,rana2023dynamite}, boxes~\cite{Xu2017,kirillov2023sam} or text~\cite{lueddecke2022clipseg,ren2024grounded}.
Out of these input types, clicks stand out for their simplicity.
Interactive segmentation approaches use them in an iterative process to refine mask outputs until accepted by the user~\cite{Xu2016,Forte2020,sofiiuk2022ritm,chen2022focalclick,rana2023dynamite}.
Input is typically provided as positive clicks on objects or negative clicks on background.
To integrate them into the network, many methods rely on a representation within clickmaps the same size as the image, which can for example be appended as additional image channels~\cite{Xu2016,Forte2020}, or passed through a separate encoder~\cite{Forte2020,sofiiuk2022ritm}.
Due to the iterative nature of interactive segmentation, networks typically further incorporate some form of structure to leverage information from previous interactions.
One way to achieve this is by employing previously existing clickmaps or mask outputs into the current iteration~\cite{Forte2020,chen2022focalclick,sofiiuk2022ritm}.

The recent advances in transformer networks have fueled the development of interactive segmentation methods that do not further rely on clickmaps.
The widely known Segment Anything Models, SAM~\cite{kirillov2023sam} and SAM2~\cite{ravi2024sam2}, perform guided segmentation through various types of prompts, including clicks.
They incorporate positional encoding for click locations and learn prompt embeddings that get cross-attended to image features.
DynaMITe~\cite{rana2023dynamite,rana2023dynamite_w} is a further transformer-based method that performs interactive segmentation without relying on clickmaps.
Instead, it leverages an architecture similar to Mask2Former, and passes click information to the network as individual queries.
This approach can solve interactive segmentation in a multi-instance manner, i.e., simultaniously for all objects in the scene.
This poses an advantage over previous methods that treat objects separately, both in terms of performance and computational efficiency.
Although a CNN-based approach and based on clickmaps, Zimmer \textit{et al.}~\cite{Zimmer2023} proposed a panoptic one-click segmentation approach within the agricultural domain, that likewise jointly handles objects in a single pass.
This is superior to handling objects separately, as it is computationally efficient and leverages knowledge about the relationship between objects.

One limitation of existing click-guided segmentation approaches is the requirement of input for every segmented object.
Related methods typically do not incorporate automatic segmentation for objects without click prompts.
Panoptic One-Click Segmentation~\cite{Zimmer2023} provides an ablation on its potential to recover from missing input clicks.
While this is an interesting first evaluation, there is no further examination of whether this could successfully be applied to semi-automate this task.
We address this by proposing the novel DropClick approach, a method for semi-automated one-click-segmentation.
DropClick is inspired by DynaMITe interactive segmentation~\cite{rana2023dynamite,rana2023dynamite_w} and Mask2Former instance segmentation~\cite{cheng2021mask2former} and segments objects in a scene both with and without user-supplied clicks.

\subsection{One-click segmentation baselines} \label{sec:related:baselines}
To the best of our knowledge, this is the first work that addresses the process of semi-automated click-based segmentation.
However, we select two related methods as baselines to compare our approach's general one-click segmentation performance, when all clicks are provided.
The first is segment anything 2 (SAM2)~\cite{ravi2024sam2} and the second being the panoptic one-click approach of~\cite{Zimmer2023}.

SAM2 is a prompt-based object segmentation method that can be used for both static images and videos, which extended on SAM~\cite{kirillov2023sam}.
It takes a number of different input types; however, we will exclusively be using the click-based prompts.
Driven by the largest segmentation dataset at the time, it is one of the first image-based foundation models that enables zero-shot segmentation.
On average, it outperforms comparable methods over a number of publicly available datasets when provided with a single click, for this reason it serves as a strong baseline to evaluate mask quality.

Finally, panoptic one-click is more of a traditional click segmentation network that was previously used on some of the data we evaluate our novel approach on.
This, coupled with their ablation study on missing click segmentation, makes it an interesting approach to evaluate against.

DropClick addresses the limitation of requiring clicks for every object in a unified transformer-based framework.
We show through extensive evaluations that our novel approach saves considerable annotation efforts in an interactive manner.
In the next section we will describe in detail the DropClick network and the process of required clicks.

\vspace{20pt}
\section{Proposed Approach} \label{sec:proposed}
\begin{figure}[t!]
\vspace{6pt}
\centering
    \includegraphics[width=.95\linewidth]{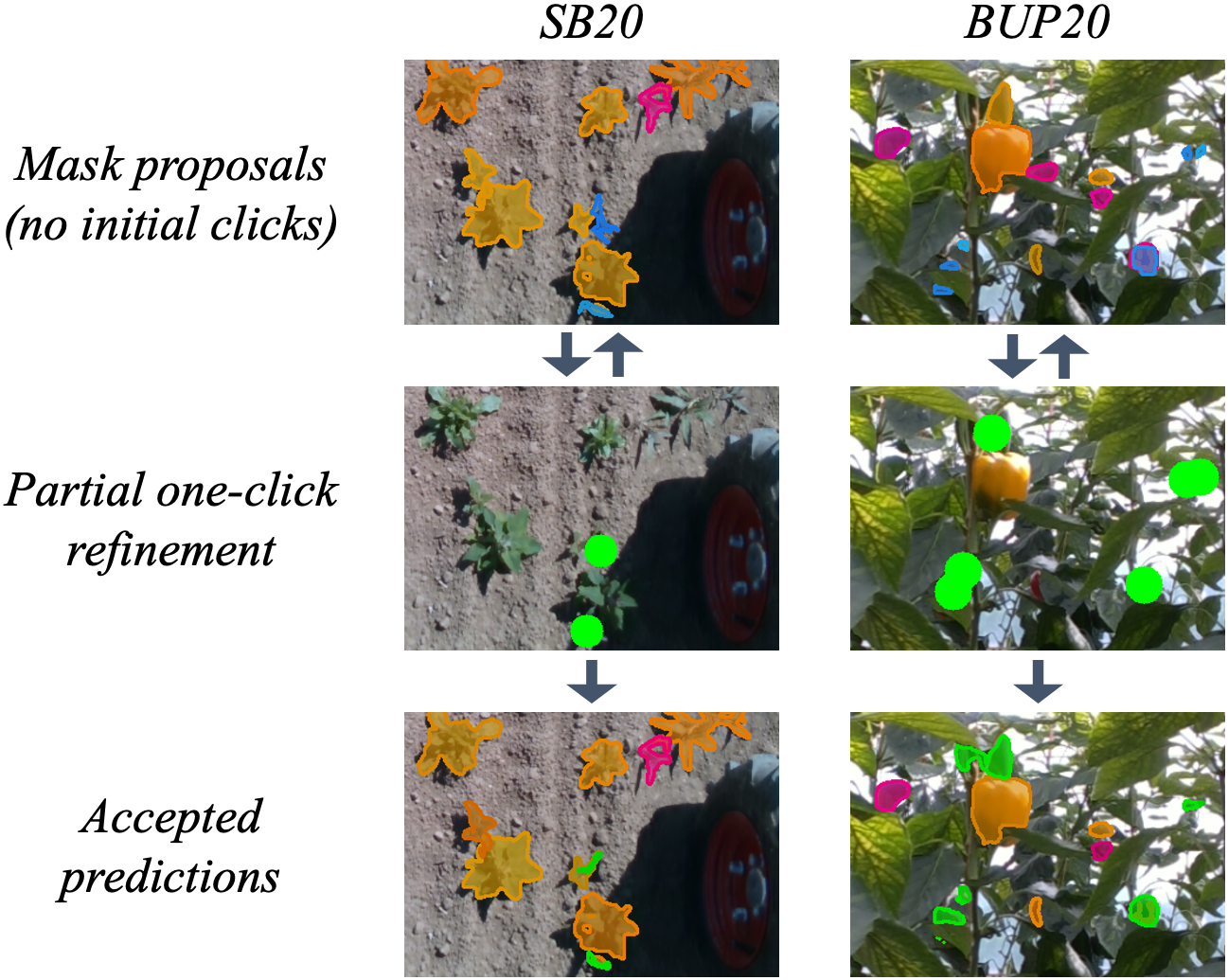}
    \caption{Clicking process for DropClick on examples from the \textit{SB20} (left column) and \textit{BUP20} (right column) datasets. Users are provided with an initial set of mask proposals (top row).
    Depicted are true positive (orange) and false positive (magenta) predictions, as well as ground truth labels for false negatives, i.e. missing predictions (blue).
    They can add an arbitrary number of one-click inputs (green dots) on objects with faulty, i.e. missing or insufficient proposals, to refine the predictions in an iterative process (middle row).
    The output from this (bottom row) are masks for both clicked (green) and non-clicked (orange) objects.
    False positive predictions can remain in the accepted result, which would typically be manually removed (after inference).}
\label{fig:clicks}
\vspace{-17pt}
\end{figure}

\begin{figure*}[t!]
    \vspace{8pt}
    \centering
    \includegraphics[width=.95\textwidth,trim={0 -2cm 0 0},clip]{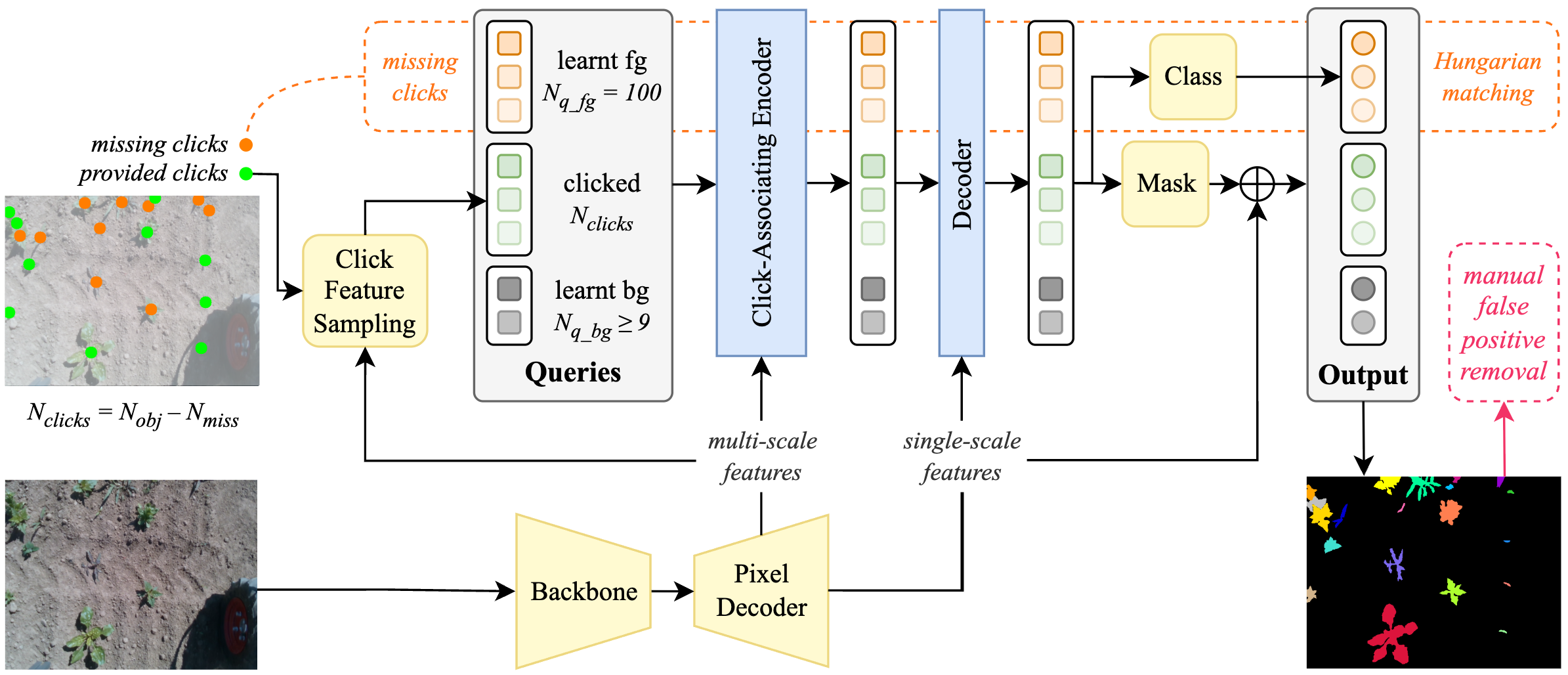}
    \caption{
    Overview of our DropClick system, which is inspired by DynaMITe~\cite{rana2023dynamite,rana2023dynamite_w} and Mask2Former~\cite{cheng2021mask2former} and solves one-click segmentation with partially missing user input.
    It takes as input an RGB image and optionally a click location per object, denoted by green circles in the top-left; objects that are not clicked are indicated by orange circles.
    Each click is passed to the network as a query, which on the output side of the network, predicts a mask for that object.
    To learn the representation for non-clicked objects, our system leverages a fixed number of $N_{q\_fg}=100$ learnt foreground queries with a classification head, and applies Hungarian matching~\cite{kuhn1955hungarian} between their output and the ground truth labels of non-clicked objects.
    Finally, DropClick produces instance-based masks for all objects in the scene, regardless of their click status.
    }
    \label{fig:scheme}
    \vspace{-10pt}
\end{figure*}

In this paper, we propose DropClick, a transformer-based, semi-automated one-click segmentation system that can be used to speed up the annotation process in agricultural datasets for robotic perception tasks.
For a given image, our system takes single clicks on objects as input to predict per-instance segmentation masks.
This output can then be used as pseudo-labels to supervise downstream tasks.
It overcomes a limitation in previous click-based segmentation methods, which typically require users to provide input for every object in the scene, in order to yield a full set of mask predictions.
To achieve this semi-automated process, DropClick combines one-click-guided and automatic instance segmentation to predict masks for both clicked and non-clicked objects at the same time.

Users are initially provided with a set of mask proposals.
They can accept these without secondary inputs, or they can refine them with further clicks on faulty predictions.
The number of clicked objects, as well as the order of clicks, is arbitrary.
However, for the best results, clicks are most important for missing or overlapping proposals.
\Cref{fig:clicks} illustrates this process on some examples from the \textit{SB20} and \textit{BUP20} datasets used in our evaluation.

Once the update process has been completed (i.e. no more missed detections), there is still the potential for false positives (FPs). 
To fully account for this in our evaluations, we include these as additional clicks, as they would generally be deleted from the predictions.
We flag these as ``additional clicks'' even though they are not input prompts and deletion would happen manually after the iterative process is completed.
This allows us to consider the fully saved number of prompts (clicks and deletions) required by the user during the annotation process.
In summary, DropClick can save a total number of user clicks $N_{saved}$ for a given image.
This is determined by the number of contained objects $N_{obj}$, subtracted by the number of clicks provided to the network $N_{clicks}$, and the number of false positives $N_{fp}$ (required deletions).
When DropClick is provided with input clicks on every object, i.e. $N_{obj}=N_{clicks}$, FPs can be automatically removed and do not require additional clicks.

\subsection{DropClick} \label{sec:proposed:dropclick}

In this section, we will provide further details on our novel one-click pseudo-label generating approach: DropClick.
Our approach is heavily inspired by DynaMITe~\cite{rana2023dynamite,rana2023dynamite_w} and Mask2Former~\cite{cheng2021mask2former}.
DynaMITe is a transformer-based approach that efficiently solves interactive segmentation simultaneously for multiple objects in a scene.
This is achieved by integrating a Mask2Former-based decoder structure with $N_{clicks}$ queries to pass $N_{clicks}$ clicks to the network.
The network learns click query embeddings from a combination of positional encodings, image features sampled from the click locations, and temporal information (to account for click sequence).
DynaMITe further integrates a set of $N_{q\_bg}$ learnt background queries, to improve foreground segmentation results by explicitly including knowledge about the image background.
These further function as a mini-batch padding approach due to the variability in click counts.
In this architecture, each output can be directly linked to its corresponding query input, as the order of queries is preserved throughout the network.
To enable one-click segmentation, we initially set the temporal click encoding to zero, essentially removing it from consideration.
We further modify the original training procedure to use only a single click per object for training the network, in contrast to multiple clicks per object.

Next, we introduce the primary contribution of this work: instance-based segmentation of non-clicked objects.
The general network architecture for our novel approach is illustrated in~\Cref{fig:scheme}.
Our network takes a fixed set of learnt foreground queries ($N_{q\_fg}=100$) similar to Mask2Former, and a classification head to generate an object score.
These queries are reserved specifically for objects that did not receive a click from the user, and are matched to the ground truth labels during training, using the Hungarian matching algorithm.
Those queries that have been sucessfully matched are labeled as an ``object'', while the remaining ones are designated ``no object''.
The classification loss is calculated only on the outputs from these $N_{q\_fg}$ learnt foreground queries, both ``object'' and ``no object'', not on those from click or background queries.
Thereby, we generate a classification score for our non-clicked objects, in addition to the existing mask output.
Alternatively, the segmentation loss is calculated on the output of all queries $N_{q\_fg}+N_{clicks}+N_{q\_bg}$.
Details about our settings for training are further described in \cref{sec:experiments:dropclick}.

At inference time, we employ the classification scores from the learnt foreground queries (missing clicks).
We concatenate the outputs from all learnt foreground queries classified as ``object'' with those from the background queries, and apply the argmax function.
The argmax function allows us to exclusively associate every pixel in the image to either one of the learnt foreground queries or background, yielding the final mask predictions for non-clicked objects.
We repeat the same process for clicked objects, leveraging the full set of outputs from the click queries while ignoring the classification scores, again in combination with those from background queries.
Finally, the combination of these post-processed outputs from click and learnt foreground queries is the full set of segmentation masks for all objects in the image, clicked or non-clicked.

\section{Experimental Setup} \label{sec:experiments}

\subsection{Datasets} \label{sec:experiments:datasets}
For all experiments we use two datasets that resemble greatly different agricultural environments, \textit{SB20} with sugar beet crops in arable farming~\cite{Ahmadi2021,Smitt2022,Halstead2021cropagnostic} and \textit{BUP20} with sweet pepper crops in a glasshouse~\cite{smitt2021pathobot}.
These datasets were captured using BonnBot-I and PATHoBot robotic platforms, respectively.
They contain instance segmentation labels for all images, and are split into 71 and 124 training, 37 and 63 validation and 35 and 93 evaluation images for \textit{SB20} and \textit{BUP20}, respectively.
Both datasets pose a number of challenges typical for agricultural environments, such as high intra-class variability, often paired with heavy occlusions.
Examples of both datasets can be seen in \Cref{fig:clicks,,fig:results:preds}.

DropClick works with small amounts of training data, as it is aimed at minimizing the overall cost of labelling robotic datasets.
We therefore hand-select $5$ of the $T$ training images from each of the original datasets and only use those to train our system.
For experiments on the downstream application of DropClick, semi-supervised training of Mask2Former, we generate pseudo-labels to replace the original annotations for the remaining $T-5$ training images.
In our experiments, when we train Mask2Former with pseudo-labels we do not use a validation set. 
This is because, in a real-world scenario the user would likely not have access to a manually annotated validation data.

\subsection{Click inputs} \label{sec:experiments:clicks}

In our analysis, we need data on click positions as input for the examined systems.
For \textit{SB20}, we use stem location labels available in the original data as described in~\cite{Zimmer2023}.
Object centers are used for \textit{BUP20}, these are derived from the original mask labels as their center of mass.
In cases where object centers lie outside the actual object, binary erosion is exploited to obtain the click position.

During training, we use a natural click randomization strategy to increase robustness against variation in user clicks.
At every epoch, we apply a 2D Gaussian-based randomization, centered at the originally calculated click locations.
Its distribution is constrained based on the object size: a larger variation for clicks within larger objects.
We limit the Gaussian's effect to a radius $R = max(a,b)/5$, where $a$ and $b$ are the side lengths the object's bounding box.
The standard deviation is defined as $\sigma=R/3$.

To evaluate the performance of DropClick under varying amounts of missing input clicks, we simulate click addition in a pre-defined order, in the same way that we instruct users to provide them.
First, we add clicks to eliminate false negatives (FNs) by arranging them in ascending order based on their IoU; a prediction is considered as a FN if the IoU threshold is $<0.4$. 
Then, when there are overlapping predictions for a non-clicked object, we attempt to resolve this by adding a click for that object; we order these in descending order of their IoU.
After resolving all these cases, the remaining objects are clicked in random order.

\subsection{Implementation and metrics}

In the following section, we provide a detailed description on our training and evaluation implementations.
For every experiment, we state how we train models to obtain the appropriate weights and further which metrics we apply to fully evaluate DropClick as a novel approach to perform click-based segmentation with partly missing user input.

\subsubsection{One-Click Segmentation Baselines} \label{sec:experiments:baseline}

We train the Panoptic One-Click Segmentation baseline according to the original settings in~\cite{Zimmer2023}.
To ensure fairness, we update the click randomization scheme and dataloader to be the same as DropClick.
For SAM2~\cite{ravi2024sam2}, we fine-tune the publicly available Hiera Large weights (v2.1 Large) for 200 epochs using the optimizer and loss configurations provided by the authors.
We measure performance using mean object IoU (mIoU) on the evaluation sets.

\subsubsection{DropClick} \label{sec:experiments:dropclick}
All DropClick experiments in this paper use a Swin-Large backbone~\cite{liu2021swin} and input transformations following~\cite{rana2023dynamite}.
The transformer architecture of our system requires pretraining, which we conduct on a combination of the 2017 COCO instance segmentation~\cite{lin2014coco} and LVIS~\cite{gupta2019lvis} datasets/annotations for 60 epochs, using a batch size of 8 and an initial learning rate of 5e-5, decayed by 0.1 at 54 and 58 epochs.
Subsequent fine-tuning on our task specific datasets, each containing only 5 images, is performed for 1000 epochs, using a batch size of 1 and a fixed learning rate of 5e-6.

To ensure DropClick can predict both clicked and non-clicked objects, images are presented to the network with a varying amount of input clicks at each iteration.
This gets determined based on probabilities $p_A=1/4$, $p_B=1/4$ and $p_C=1/2$, where $p_A$ corresponds to dropping no clicks, $p_B$ to dropping all $N_{obj}$ clicks and $p_C$ to dropping a random number of clicks between $1$ and $N_{obj}-1$.

We compute losses as seen in~\cite{cheng2021mask2former}, binary cross entropy with sigmoid $\mathcal{L}_{bce}$ and dice $\mathcal{L}_{dice}$ for the mask head and cross entropy $\mathcal{L}_{cls}$ for classification.
However, we further separate mask loss computation for clicked and learnt queries in order to vary their weights $w$ such that $w_{bce\_clicked} = 5$ and $w_{dice\_clicked} = 5$, while $w_{bce\_learnt} = 10$ and $w_{dice\_learnt} = 25$.
Classification loss is exclusive for learnt queries with weight $w_{cls} = 10$.

As a general measure for DropClick's mask quality, we use mIoU performance over the evaluation sets.
We further conduct detailed analysis on its ability to perform under missing input clicks.
We do so by reporting the number of false negative (FN) and false positive (FP) detections, as well as mIoU under gradually increasing amounts of object clicks which are added as described in~\Cref{sec:experiments:clicks}.
We first analyze the evaluation sets, to identify the ratio of clicks $r_{miss}$ that can reliably be dropped while maintaining high performance in terms of mIoU and low FN and FP counts.
When choosing a fixed $r_{miss}$, it needs to be considered that the exact number of missing clicks $N_{miss}$ in an image is naturally dependent on its total number of contained objects $N_{obj}$.
We derive it as $N_{miss}=\lceil{r_{miss}}*N_{obj}\rceil$, i.e., the smallest integer greater than or equal to the product of $r_{miss}$ and $N_{obj}$.
Therefore, a dataset's final $r_{miss}$ is likely to be slightly higher than the initially chosen value.
This effect can be quite large and undesirable in datasets containing images with very low $N_{obj}$, however, this is not the case in our data.

\subsubsection{Semi-supervised instance segmentation}
We use DropClick generated pseudo-labels within our training data as described in~\Cref{sec:experiments:datasets} to train Mask2Former instance segmentation~\cite{cheng2021mask2former} in a semi-supervised manner.
We initialize models with a Swin-Large backbone~\cite{liu2021swin}, using the original weights and configuration provided by the authors.
We fine-tune our model for 50 epochs, which we identified as the convergence point for both datasets for fully-supervised training.
We apply the same number of training epochs within the other experiments, where validation data is missing.
Finally, we report performance as segmentation AP50 over the evaluation sets.

\vspace{-1pt}
\section{Results} \label{sec:results}

\begin{table}[!b]
    \vspace{-6pt}
        \caption{One-click segmentation performance (mean object IoU) on \textit{SB20} and \textit{BUP20} evaluation sets, for baselines and DropClick. All models are trained on small subsets of 5 images from the original training sets. DropClick results are from multiple evaluations, each with a different amount of input clicks provided to the network (clicked on all objects, \mytexttilde50\% missing and no clicks provided).}
        \label{tab:results:weakseg}
        \centering
            \begin{tabular*}{0.9\columnwidth}{l@{\extracolsep{\fill}}rr}
            \toprule
            \textit{Method} & \textit{SB20} & \textit{BUP20} \\
            \midrule
            SAM2 (v2.1 Large)~\cite{ravi2024sam2} & 67.4 & 71.1 \\
            Panoptic One-Click~\cite{Zimmer2023} & \textbf{71.9} & 58.6 \\
            \cmidrule(lr){1-3}
            DropClick (all clicks provided) & 70.0 & \textbf{72.6} \\
            DropClick (\mytexttilde50\% missing) & 68.9 & 71.3 \\
            DropClick (no clicks provided) & 61.8 & 58.3 \\
            \bottomrule
            \end{tabular*}
\end{table}

We perform three sets of experiments to demonstrate the viability of DropClick.
First, we compare DropClick's mask quality against two baseline systems, SAM2~\cite{ravi2024sam2} and Panoptic One-Click Segmentation~\cite{Zimmer2023}.
Second, we examine how DropClick performs as we vary the number of missing clicks. 
This demonstrates the strength of DropClick highlighting that considerably less user input is needed to still obtain high performance.
Third, we verify the applicability of DropClick pseudo-labels for the downstream task of semi-supervised instance segmentation.

\subsection{One-click segmentation}

\begin{figure}[tp]
\vspace{5pt}
    \centering
    \includegraphics[width=.9\linewidth]{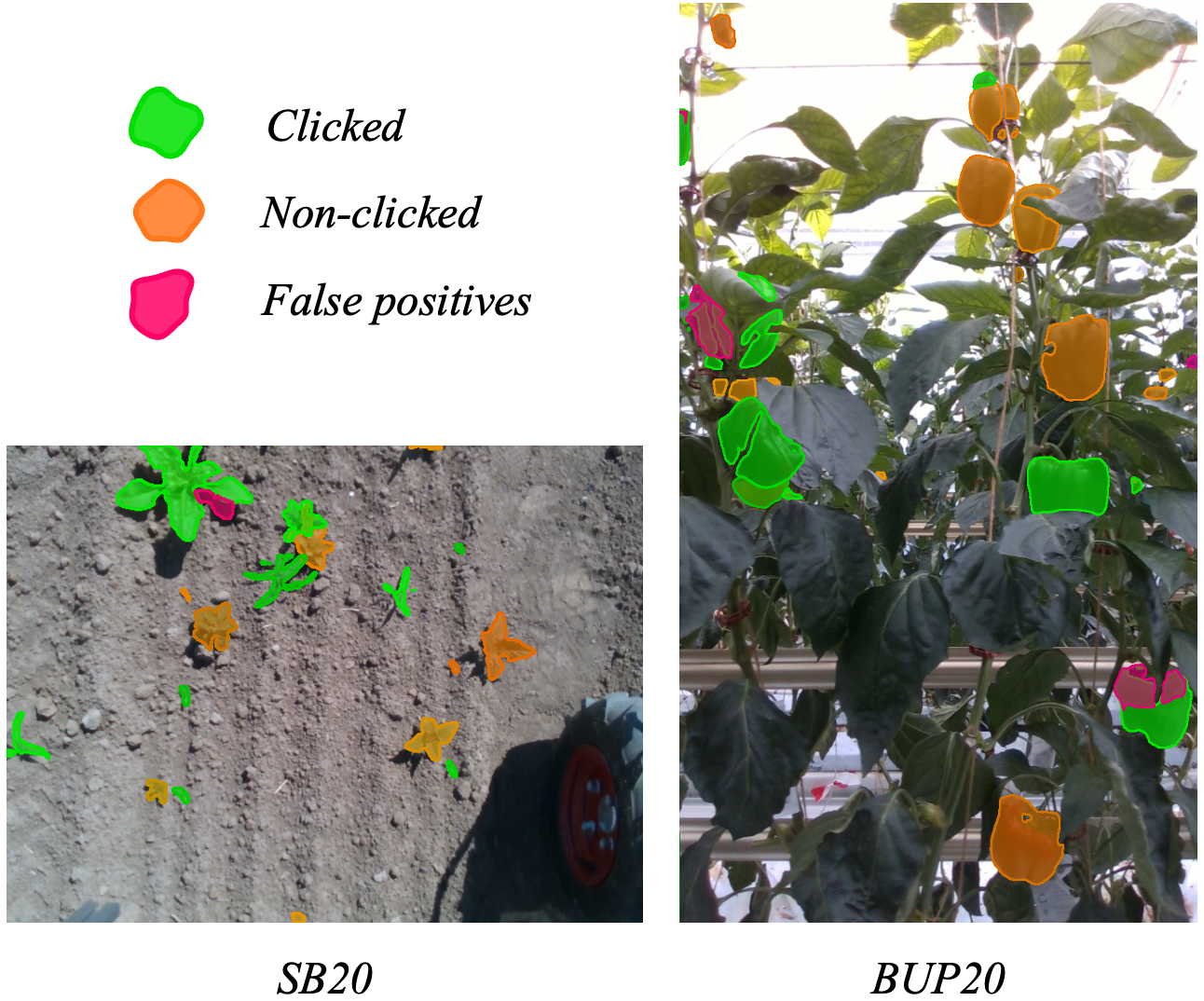}
    \caption{Example predictions from DropClick models on selected \textit{SB20} (left) and \textit{BUP20} (right) evaluation images at approximately 50\% of input clicks missing.
    Orange-colored masks show true positive predictions that did not receive any input, thus resembling free pseudo-labels.
    Green ones are associated to such objects that have each received a single input click.
    Magenta marks false positive predictions which require removal after inference by the user, thereby adding to the total input cost by one additional click each.
    }
    \label{fig:results:preds}
\vspace{-12pt}
\end{figure}

The results in~\Cref{tab:results:weakseg} demonstrate that DropClick provides consistently high segmentation performance across both datasets.
This is achieved both when all clicks are supplied and when 50\% are missing.
With fully click-annotated images, DropClick outperforms SAM2 with an absolute improvement of 2.6 and 1.5 points for mIoU on \textit{SB20} and \textit{BUP20}, respectively.
In comparison to Panoptic One-Click, DropClick has slightly lower performance on \textit{SB20}, with an absolute decrease of 1.9 points for mIoU.
However, for \textit{BUP20} we report considerably higher performance with an absolute increase of 14.0 points for mIoU.
This demonstrates that DropClick provides consistently high-quality pseudo-labels across agricultural domains when all user clicks are provided.

The power of DropClick is most evident when we only click on approximately 50\% of the objects; a feature unique to DropClick.
As we reduce the number of supplied clicks (50\%), we only observe a minor degradation in performance.
For \textit{SB20} we decrease performance by only 1.1 points, and for \textit{BUP20} by only 1.3 points, which still outperforms SAM2 even though half of the clicks are missing.
When all input is removed, performance decreases by 8.2 points for \textit{SB20} and 14.3 points for \textit{BUP20}.

Overall, these results show that DropClick is generally able to provide mask quality on a high level, comparable to or better than similar approaches.
More importantly, they support the main strength of our system, which is to perform on the same level when users only provide partial input.

\begin{table}[!b]
    \vspace{-6pt}
    \caption{DropClick mean object IoU performance and final click savings (average numbers per image) when \mytexttilde50\% of initial object clicks are removed during pseudo-label generation on the training sets (excluding those 5 images used to train DropClick).}
    \label{tab:results:savings}
    \centering
    \begin{tabular*}{0.9\columnwidth}{l@{\extracolsep{\fill}}rr}
    \toprule
    \                            & \textit{SB20}       & \textit{BUP20} \\
    \midrule
    mIoU                         & 69.0                & 70.0           \\
    \cmidrule(lr){1-3}
    $N$ objects                  & 19.0                & 28.5           \\
    $N$ objects clicked          & 8.7                 & 13.5           \\
    $N$ false positives          & 1.5                 & 5.9            \\
    $N$ clicks saved             & 8.8                 & 9.1            \\
    \cmidrule(lr){1-3}
    \% clicks saved              & \textbf{46.3}       & \textbf{31.9}  \\
    \bottomrule
    \end{tabular*}
\end{table}

\subsection{Removing Input Clicks}\label{sec:results:savings}

\begin{figure}[tp]
\vspace{6pt}
    \centering
    \includegraphics[width=.98\linewidth]{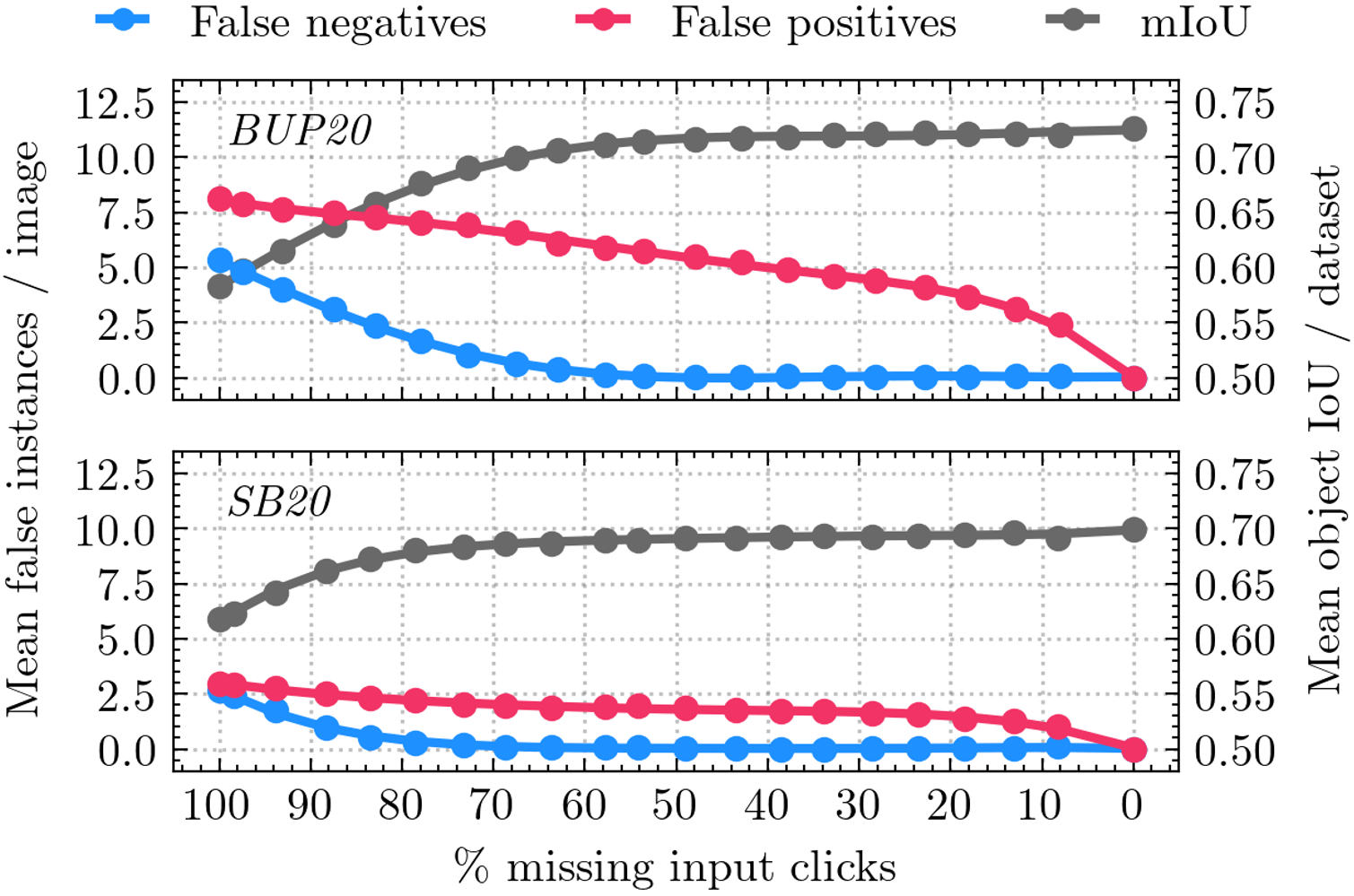}
    \caption{DropClick performance on \textit{SB20} (bottom) and \textit{BUP20} (top) evaluation sets at different ratios of missing input clicks.
    Shown are mean object IoU (mIoU, gray) over the tested data, as well as mean number of false instances per image as false negatives (blue) and false positives (magenta).
    False positives add to the total user input cost, as they require an additional click each to be removed (after inference).
    At 0\% missing input clicks, FPs can be automatically removed, hence they are set to zero.
    }
    \label{fig:results:missing}
\vspace{-10pt}
\end{figure}

In this experiment, we examine the performance of DropClick as we vary the number of input clicks.
\Cref{fig:results:missing} shows the results on the evaluation sets of \textit{SB20} (bottom) and \textit{BUP20} (top).
From left to right, the input clicks are gradually increased in increments of $r_{miss}=5\%$ from 100\% missing clicks to no missing clicks.
The discussed results are based on the mean object IoU (mIoU), average number of false negatives (FN) and false positives (FP) per image.

Our results clearly show that DropClick does not need the full set of input clicks to reach high performance.
As expected, all metrics improve as the percentage of missing clicks is decreased.
For both mIoU and FNs we see that a saturation point in the results is achieved as early as 70\% (\textit{SB20}) and 50\% (\textit{BUP20}) of clicks are still missing.
Based on this observation, we selected 50\% as our critical number of missing clicks and applied it to our remaining experiments, as this is a reliable point across both datasets.
The final component of this evaluation is the FPs.
At 50\% missing clicks DropClick achieves 1.5 and 5.9 FPs per image from \textit{SB20} and \textit{BUP20}, respectively.
\Cref{fig:results:preds} outlines prediction examples, including the FPs witnessed during our experiments.
However, even with these anomalies, DropClick yields high-quality masks for both clicked and non-clicked objects in the scene.

Here, we additionally examine the final amount of click savings that DropClick achieves in our following experiment on semi-supervised instance segmentation.
As both 50\% of object clicks as well as clicks to remove FPs are provided, \Cref{tab:results:savings} summarises these results and shows that large amounts of clicks are spared when creating the pseudo-labeled subsets of the original training sets.
Combining the number of both missing clicks and FPs, we achieve final click savings of 46.3\% and 31.9\%, for \textit{SB20} and \textit{BUP20} respectively.
At the same time, we observe high pseudo-label quality of 69.0 and 70.0 points mIoU, comparable to that from the evaluation sets.

\subsection{Semi-supervised Instance Segmentation}\label{sec:results:semi}

\begin{table}[!b]
    \vspace{-6pt}
    \caption{Mask2Former instance segmentation performance (AP50) trained
    (1) fully supervised using the original training data, 
    (2) semi-supervised by DropClick pseudo-labels with a single click per object, 
    (3) semi-supervised by DropClick pseudo-labels with partially missing clicks and 
    (4) only on the same 5-image subsets of data used to train DropClick.
    }
    \label{tab:results:semi}
    \centering
    \begin{tabular*}{0.9\columnwidth}{l@{\extracolsep{\fill}}rr}
    \toprule
    \textit{Supervision level}              & \textit{SB20}  & \textit{BUP20} \\
    \midrule
    Fully supervised	                    & \textbf{72.0}	 & \textbf{80.1} \\
    \cmidrule(lr){1-3}
    Semi-supervised (all clicked)	        & 70.7           & 77.0 \\
    Semi-supervised	(partially missing)     & 70.1	         & 77.0 \\
    \cmidrule(lr){1-3}
    Subset only (5 images)	                & 61.6	         & 65.5 \\
    \bottomrule
    \end{tabular*}
\end{table}

The final evaluation uses DropClick to create pseudo-labels for semi-supervised training of Mask2Former.
Our results confirm that DropClick is a suitable method for this downstream task, even when 50\% of its user input is missing.
From~\Cref{tab:results:semi} we see that our baseline, fully supervised on both training sets from the respective datasets, achieves AP50 results of 72.0 and 80.1 for \textit{SB20} and \textit{BUP20}, respectively.
When just using the 5 fully labeled images used for training DropClick (Subset only) we, as expected, observe a drastic degradation in results by 10.4 and 14.6.
In contrast, we observe a much smaller decrease by only 1.3 points AP50 for \textit{SB20} and 3.1 for \textit{BUP20}, for our semi-supervised models with full click inputs.
Finally, our models based on 50\% missing clicks perform on the same level with a negligible further decrease of 0.6 points for \textit{SB20} and no difference for \textit{BUP20}.
These results confirm that DropClick is a suitable method for pseudo-label generation for semi-supervised learning of agricultural robotics data, with only a slight decrease in accuracy for a considerable increase in efficiency.

\section{Conclusion} \label{sec:conclusion}

In this paper, we have presented DropClick, a novel method for accurate and efficient annotation of segmentation datasets for agricultural robotics.
Operating on user input in the form of single clicks per object, our system can vastly reduce the costs of this task.
We have demonstrated that DropClick produces high quality mask outputs, even when 50\% of its click input is removed.
We successfully applied pseudo-labels generated through our system to train Mask2Former in a semi-supervised manner, for instance segmentation of plants and sweet peppers.
In this application, partly removing input clicks from DropClick maintained high performance, yielding 70.1 points AP50 compared to 70.7 for \textit{SB20} data and no difference between the two models at 77.0 for \textit{BUP20}.
Thereby, we were able to save 46.3\% and 31.9\% of total input for \textit{SB20} and \textit{BUP20}, respectively.

\section*{Acknowledgements}
This work was partly funded by the Deutsche Forschungsgemeinschaft (DFG, German Research Foundation) under Germany’s Excellence Strategy - EXC 2070 – 390732324 and partly funded by the German Federal Ministry of Food and Agriculture (BMEL) through the WeedAI project.

\bibliography{references}

\end{document}